\documentclass[journal]{IEEEtran}
\usepackage{cite}

\usepackage{multirow}
\usepackage{booktabs}

\ifCLASSINFOpdf
  \usepackage[pdftex]{graphicx}
  \usepackage{caption}
  \usepackage{subcaption}
  \usepackage{wrapfig}
  \usepackage{multicol}
  \graphicspath{{../pdf/}{../images/}}
\else
\fi
\usepackage{amssymb}
\usepackage{amsmath}
\usepackage{algorithm,algpseudocode}

\algnewcommand{\Inputs}[1]{%
  \State \textbf{Inputs:}
  \Statex \hspace*{\algorithmicindent}\parbox[t]{1.0\linewidth}{\raggedright #1}
}

\algnewcommand{\Init}[1]{%
  \State \textbf{Initialization:}
  \Statex \hspace*{\algorithmicindent}\parbox[t]{.8\linewidth}{\raggedright #1}
}

\algnewcommand{\Initialization}[1]{%
  \State \textbf{Initialization of the Kalman Filter:}
  \Statex \hspace*{\algorithmicindent}\parbox[t]{.8\linewidth}{\raggedright #1}
}

\algnewcommand{\Variables}[1]{%
  \State \textbf{Variables:}
  \Statex \hspace*{\algorithmicindent}\parbox[t]{.8\linewidth}{\raggedright #1}
}

\algnewcommand{\Outputs}[1]{%
  \State \textbf{Outputs:}
  \Statex \hspace*{\algorithmicindent}\parbox[t]{.8\linewidth}{\raggedright #1}
}

\usepackage{url}

\usepackage[symbol]{footmisc}

\begin{document}
%
\title{Monocular visual autonomous landing system for quadcopter drones using software in the loop}
%
%
%

\author{
    \IEEEauthorblockN{Miguel~Saavedra-Ruiz\IEEEauthorrefmark{1}, Ana Maria~Pinto-Vargas\IEEEauthorrefmark{2} \thanks{Miguel~Saavedra-Ruiz and Ana Maria~Pinto-Vargas undertook this work while they were affiliated to Universidad Autónoma de Occidente}, Victor~Romero-Cano,~\IEEEmembership{Member,~IEEE} \IEEEauthorrefmark{3}}
    \\\IEEEauthorblockA{\IEEEauthorrefmark{1}Mila - Quebec Institute of Artificial Intelligence, Université de Montréal, Canada
    \\\ miguel.angel.saavedra.ruiz@umontreal.ca}
    \\\ \IEEEauthorblockA{\IEEEauthorrefmark{2} Alternova Tech SAS, Medellín, Colombia
    \\\ ana.pinto@alternova.io}
    \\\ \IEEEauthorblockA{\IEEEauthorrefmark{3}Robotics and Autonomous Systems Laboratory, Faculty of Engineering, Universidad Autónoma de Occidente, Cali, Colombia
    \\\ varomero@uao.edu.co}
}

\maketitle

\begin{abstract}
Autonomous landing is a capability that is essential to achieve the full potential of multi-rotor drones in many social and industrial applications. The implementation and testing of this capability on physical platforms is risky and resource-intensive; hence, in order to ensure both a sound design process and a safe deployment, simulations are required before implementing a physical prototype. This paper presents the development of a monocular visual system, using a software-in-the-loop methodology, that autonomously and efficiently lands a quadcopter drone on a predefined landing pad, thus reducing the risks of the physical testing stage. In addition to ensuring that the  autonomous landing system as a whole fulfils the design requirements using a Gazebo-based simulation, our approach provides a tool for safe parameter tuning and design testing prior to physical implementation. Finally, the proposed monocular vision-only approach to landing pad tracking made it possible to effectively implement the system in an F450 quadcopter drone with the standard computational capabilities of an Odroid XU4 embedded processor. 
\end{abstract}

\begin{IEEEkeywords}
Autonomous landing, quadcopter, target tracking, software-in-the-loop, simulation, Sim2Real.
\end{IEEEkeywords}

%
\IEEEpeerreviewmaketitle

\section{Introduction}
%
%
%
%
\IEEEPARstart{U}{nmanned} Aerial Vehicles (UAVs) have recently become popular due to their potential in terms of performing complex tasks such as infrastructure inspection \cite{a}, target  detection \cite{b,c}, or search and rescue \cite{d}. The use of these gadgets has led to both substantial improvements in the efficiency of these processes and a reduction in human casualties while performing hazardous labours. The deployment of UAVs in such applications requires a complete suite of sensors such as GPS, laser rangefinders, radar and cameras \cite{e}, which can be used to endow the vehicle with environmental awareness and the capability to perceive events of interest. However, the use of many peripherals in a UAV requires an extensive amount of on-board computational resources and power that are not always available owing to the vehicle’s dimensions and the high implementation costs.

\begin{figure*}[ht]
  \centering
  \includegraphics[width=1\textwidth,keepaspectratio]{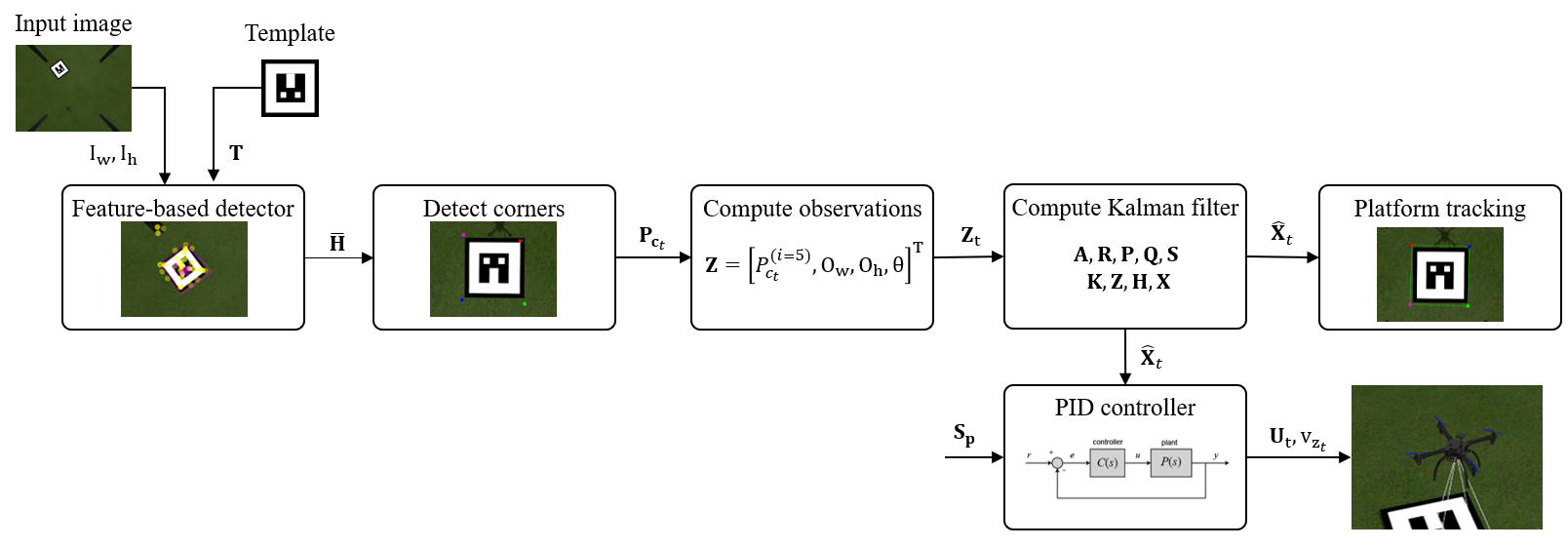}
  \caption{Autonomous landing system pipeline with image-based visual servoing.}
  \label{fig:f1}
\end{figure*}

Cameras have been proposed as a feasible alternative to overcome these issues, as they have a relative low price and enable the estimation of content-rich representations of the environment. For instance, cameras have been widely employed in various tasks such as mapping \cite{f} and object tracking \cite{g}. Further research efforts have been conducted to make use of cameras in the development of visual-based autonomous landing systems for UAVs. Autonomous landing maneuvers remain a crucial task for rotorcraft, and allow the development of complete, end-to-end autonomous flying vehicles that are capable of performing complex assignments such as those mentioned above.

Most state-of-the-art visual-based landing systems have shown unprecedented results that are comparable to the performance of UAVs with a full suite of sensors \cite{h}.
The employment of natural landmarks to land a rotorcraft in unstructured environments is a strategy commonly used for emergency landing situation \cite{1h, f}. These methods rely on the use of vision-based SLAM algorithms such as ORB\_SLAM2 \cite{2h} for localization of the vehicle and mapping of appropriate landing spots in the environment. Nevertheless, these techniques are prone to deliver low spatial resolution and be computationally expensive, hindering the performance of autonomous landing applications.

On the other hand, the utilization of artificial landmarks is one of the most traditional techniques used in landings on both static \cite{i} and moving platforms \cite{j}. The extraction of information from a landmark, such as the relative pose or template coordinates, is broadly used in the application of image-based visual servoing (IBVS), a technique that performs the majority of the control calculations in 2D image space \cite{k} and reduces the computational load of a small rotorcraft while landing. 

Visual servoing is commonly used with classical computer vision methods to track features over several image frames and to create stable control references to land the aircraft on a desired target. Despite the potential of IBVS in terms of autonomous landing, additional assumptions are required, for example that the features in the image are static features of the object, or that the object does not leave the field of view \cite{g}. Furthermore, the implementation of visual-based autonomous landing systems requires rigorous assessments in simulated environments to identify possible perils before deploying the whole system in the real world.

Lately, deep learning has been proposed as an alternative to replace feature-based methods with Convolutional Neural Networks (CNNs) for landmark detection \cite{1g}. The use of CNN has exhibited robustness with diverse lighting conditions, scale variations and rotations. Notwithstanding the potential of deep learning-based object detectors, these models typically require extensive amount of human-labeled datasets and vast computational resources that are usually available only with off-board computing strategies \cite{2g}.

In this work, we address these problems by proposing a complete monocular visual-based perception and control strategy for the autonomous landing of a UAV in a Gazebo-based simulated environment. This system aims to mitigate the current limitations on classic computer vision-based methods created by changes in the appearance in the image by using a Kalman filter to estimate the position of the template throughout the landing process. Additionally, the use of only IBVS techniques for the control of the aircraft reduces the computational cost of the system and eliminates the need for expensive 3D position reconstruction calculations, thus allowing for real-time control of small UAVs with low-cost computers.


Fig.  \ref{fig:f1}  illustrates  the  general  workflow  of  the  proposed method. Initially, the system computes the homography matrix between the current image frame and the predefined template, using  a  feature-based  detector.  Next,  the  homography  matrix is used to compute the corners and the centroid of the object in  the  current  image  frame.  These  points  are  then  passed to  a  Kalman filter  estimation  module.  Finally,  the  Kalman filter  estimations  are  used  to  track  the  template  in  the  image frame, and as a process variable for a set of three PID-based controllers that perform the safe landing of the vehicle. 

The full system was developed and assessed in a Gazebo-based simulated environment in order to bridge the gap between real-world deployment and theory, and to reduce the number of risks while the vehicle is tested. All the parameters for the vision and control systems employed in the Gazebo-based simulation were directly transferred to the real-world quad-rotor in a zero-shot\footnote[2]{It refers to when parameters are learned or set in a source domain (simulation) and tested without \textit{fine-tuning} in a target domain (real-world).} sim2real (simulation to reality) fashion in order to validate that these simple approaches can be effectively transferred to the vehicle without additional tuning \cite{1k}. Overall, the principal contributions of this work can be summarized as follows: 

\begin{enumerate}
  \item A complete, flexible, Gazebo-based simulation of a visual-based landing system for low-cost UAVs;
  \item The implementation of a Kalman-filter-based methodology for landing platform tracking using monocular vision in both a simulated and a physical drone;
  \item A control strategy for quadcopter landing that is seamlessly implemented using the popular PX4 software-in-the-loop (SITL) Gazebo interface, which facilitates its transfer to a physical drone.
\end{enumerate}


This paper consists of five sections, as follows: Section II presents related work. In Section III, the feature-based detector and Kalman filter are explained. Section IV describes the proposed control strategy for the landing maneuver, while Sections V and VI contain the simulated and experimental results, respectively. Finally, Section VII presents the conclusion.

\section{Related Work}

Autonomous landing for multi-rotor aircraft is a problem that has been extensively studied. Various approaches have relied on the use of vision-based techniques to identify the salient features in an image and to land the vehicle on both static \cite{l, m, n} and moving platforms \cite{h,j,k}. Classic computer vision methods, such as feature-based extraction and description or homography-based approaches \cite{o, p}, are commonly used to estimate the relative pose of the vehicle with respect to a landing platform at a relative low computational cost. 



Spatial information can be extracted from natural and artificial landmarks. In \cite{f}, the authors proposed the use of natural landmarks for the detection and reconstruction of landing sites based on the texture of the ground. Visual-based SLAM techniques are also exploited to assemble world's representations and find feasible landing spots for the rotor-craft in unstructured environments as shown in \cite{1h}. Similarly, the use of artificial landmarks can alleviate the autonomous landing task by providing references with known dimensions for detection and tracking over several image frames \cite{r}.

The use of markers has been exploited to provide a traceable reference for landing control systems and to enhance the position estimation of aerial vehicles through visual inertial odometry (VIO). In \cite{g}, the authors estimated the relative pose of the aircraft with respect to a spherical target and used an extended Kalman filter to fuse these measurements with IMU data to accurately locate the vehicle within the space. 

Kalman filters are not exclusively employed to fuse information from multiple sensor sources but also to estimate the states of a system from a unique noisy source \cite{s, 1s}. These estimations are used in IBVS, with linear control strategies such as nested PIDs \cite{f, l} and nonlinear ones like sliding mode controllers \cite{k}, to accurately land an aerial vehicle. The utilization of Gaussian estimators for IBVS provides numerically stable and continuous references for controllers, even when the object of interest is outside the field of view of the camera.

Further research efforts have concentrated on the use of deep learning methods to detect and track landmarks in images using CNN-based architectures \cite{2g, t, u} or to automate the complete landing task with deep reinforcement learning (DRL) agents \cite{v}. However, the use of artificial neural networks requires substantial computational resources for real-time inference and thousand of human-labeled images based on the task at hand \cite{2g}. Likewise, visual-based 3D reconstruction techniques tend to be computationally expensive for on-board computers in small UAVs \cite{1h}, and need to satisfy various assumptions to achieve accurate pose estimations.

We aim to reduce the computational load when performing IBVS with the use of a vision-based tracking system, and to produce a stable reference for a set of nested PID-based controllers similar to those in \cite{f}. The idea behind the detection and tracking system is to produce a 2D image-based reference for the controller, thus avoiding expensive 3D pose reconstructions as in \cite{k}. In this work, the use of the Kalman filter is restricted to filtering 2D estimations of the landing pad from noisy observations, unlike the application of VIO in most other related work. Contrary to commonly used simulation tools like \textit{RotorS} \cite{2k}, which provide Gazebo-based simulation environments for multi-rotor drones with no interface with a real flight controller, our implementation utilises the SITL provided by PX4, which runs the Pixhawk flight stack, and therefore provides direct support to the physical robot deployment process.

\section{Vision system}


This section describes our detection and tracking system for the autonomous landing of a UAV, which is an extension of our previous work in \cite{o, y}. We first explain how the feature-based object detector detects the landing platform when comparing the platform’s template with the input image. Next, we describe how the system translates the corners and the centroid of the detected platform from the homography matrix to a vector that contains the system observations. Finally, we explain how a tailored Kalman filter is used to estimate the pose of the landing platform, even when no detection has been obtained.

\subsection{Feature-based object detection}
\label{section:fbod}

Object detection is a crucial task in robotic perception. Feature-based detectors and descriptors are widely used, due to their speed in computing the salient features of images. For increased robustness in object detection, these features should be invariant to rotation, scale and affine transformations over several frames \cite{w}. To find correspondences between two images, we consider a set of features in the template image \(\boldsymbol{F_T} \in \mathbb{R}^n\) and the current frame \(\boldsymbol{F_S} \in \mathbb{R}^m\), where \(n, m \in \mathbb{Z}\) represent the number of features in each image. Each feature in the template and scene frames is associated with a descriptor \(\boldsymbol{D_T} \in \mathbb{R}^{n \times k}, \boldsymbol{D_S} \in \mathbb{R}^{m \times k} \), where \(k\) is the dimension of the descriptor for each feature.

\begin{figure}[ht]
    \centering
    \includegraphics[width=0.4\textwidth,height=0.4\textheight,keepaspectratio]{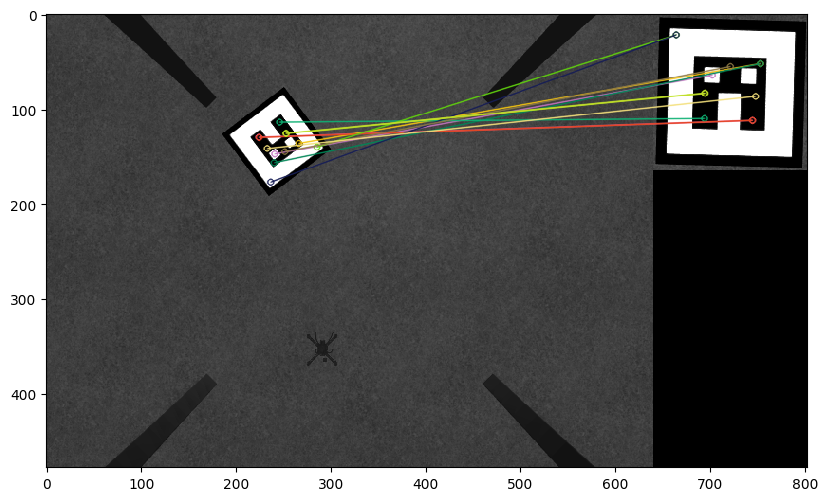}
    \captionsetup{justification=centering}
    \caption{Feature matching between the features of the template \(\boldsymbol{F_T}\) and the features of the scene image \(\boldsymbol{F_S}\)}
    \label{fig:f2}
\end{figure}

With a set of descriptors, it is possible to compute matches between image pairs by performing distance calculations, such as the Euclidean distance between the descriptors of the template and the scene, as shown in \eqref{eq:1}. Two features are matched when the closest descriptors between two images in the descriptor space have been found. As a result, similar features in the template image \((x_i, y_i)\) are matched with the similar pair \((x_i', y_i')\) in the current image frame, as illustrated in Fig. \ref{fig:f2}. 

\begin{equation}
    \label{eq:1}
    \boldsymbol{d}_i = min(\sum_{i = 1}^n\sum_{j = 1}^m \sqrt{(\boldsymbol{D_T}^{(i,k)})^2 - (\boldsymbol{D_S}^{(j,k)})^2})
\end{equation}

\subsubsection{Homography Matrix}

Finding correspondences between image pairs allows us to compute the homography matrix \(\boldsymbol{\overline{H}} \in \mathbb{R}^{3 \times 3}\). This matrix is a transform that maps points from one image frame (template) to the corresponding points in the other image frame (scene). To compute the homography, at least four matches are needed. Then, knowing the homography between two images and the dimensions of the template \(\boldsymbol{T} = [w_T, h_T]^T\), it is possible to apply a perspective transform that maps the template position from the template image to the scene image using \eqref{eq:2}.

\begin{equation}
    \label{eq:2}
    \begin{bmatrix} x^{'} \\ y^{'} \\ 1 \end{bmatrix} = \boldsymbol{\overline{H}} \begin{bmatrix} x \\ y \\ 1 \end{bmatrix} = \begin{bmatrix} h_{11} & h_{12} & h_{13} \\ h_{21} & h_{22} & h_{23} \\ h_{31} & h_{32} & h_{33} \end{bmatrix} \begin{bmatrix} x \\ y \\ 1 \end{bmatrix}
\end{equation}

In \eqref{eq:2}, \([x, y, 1]^T\) are the coordinates of points (e.g. corners) in the template image and \([x', y', 1]^T\) are the same points mapped in the scene image, where \(\boldsymbol{\overline{H}} : \mathbb{R}^3 \rightarrow \mathbb{R}^3\). Object detection with feature-based methods and homography calculations tends to speed up the process and can provide a reliable estimation of the location of the object of interest in the current image frame.

\subsection{System observations}

Using the homography matrix, the corners and centroid of the template detected in the current image frame can be computed. \(\boldsymbol{P_c}_t \in \mathbb{R}^{5 \times 2}\) is defined as a vector of coordinates, where each row corresponds to a \(x,y\) point at time index \(t\).

\begin{figure}[ht]
    \centering
    \includegraphics[width=0.25\textwidth,height=0.25\textheight,keepaspectratio]{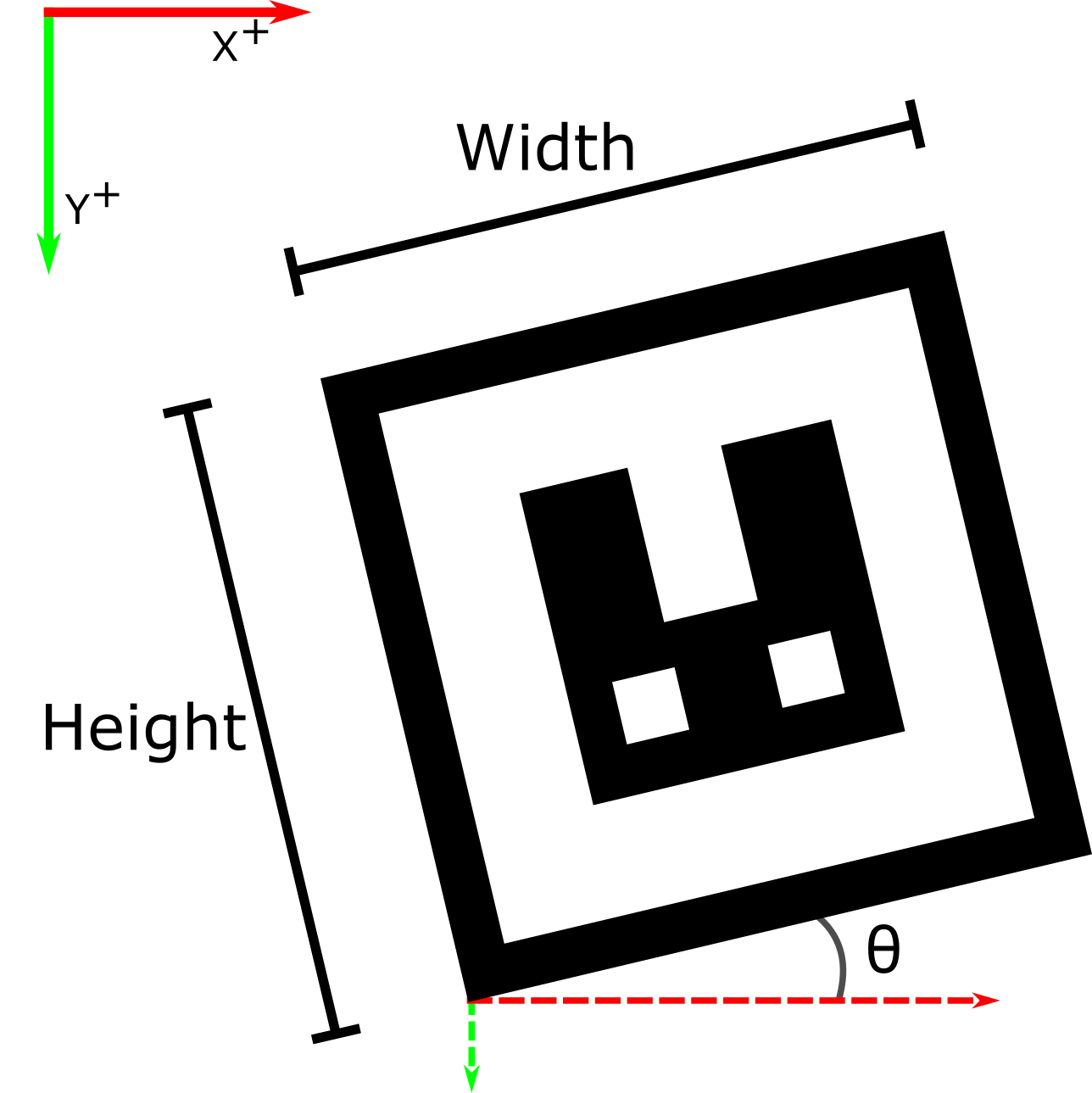}
    \captionsetup{justification=centering}
    \caption{System's observations and coordinate system. The vector  \(\boldsymbol{Z}_t\) is created at each time-step based on the width and height (\(O_w, O_h\)), the orientation \(\theta\) and the position of the centroid (\(x,y\)) of the template in the current image frame.}
    \label{fig:obs_states}
\end{figure}

These points are used to determine the observations that will be fed into the Kalman filter. The vector of observations at time \(t\) is defined as \(\boldsymbol{Z}_t = [\boldsymbol{P_c}_t^{(i = 5)}, O_w, O_h, \theta]\), where \(\boldsymbol{P_c}_t^{(i = 5)}\) are the \(x,y\) centroid coordinates of the landing pad; \(O_w, O_h\) are the width and height of the template, respectively; and \(\theta\) represents the angle of the template with respect to the \(x\) axis of the image, as shown in Fig. \ref{fig:obs_states}. 

\subsection{Kalman filter}

The Kalman filter is an estimator that infers hidden states from indirect, inaccurate and uncertain observations. It is possible to use the Filter to handle noisy observations from the detection module and produce a continuous estimate of the template position at each time step \(t\) \cite{f}.

We assume that we have a Linear Dynamic System (LDS) for a landing platform such as in \eqref{eq:3}, where \(x_t\) is the \(x\) coordinate of a pixel at time index \(t\) and \(\Delta t\) is the time between two consecutive image frames. Similarly, in \eqref{eq:4}, \(\Dot{x}_{t}\) corresponds to the \(x\) velocity component of a pixel in the image.

\begin{equation}
    \label{eq:3}
    x_t = x_{t-1} + \Delta t \Dot{x}_{t-1}
\end{equation}

\begin{equation}
    \label{eq:4}
    \Dot{x}_{t} = \Dot{x}_{t-1}
\end{equation}

The set of states \(\boldsymbol{X} \in \mathbb{R}^{10}\) is given by \eqref{eq:5}, with \(x_c, y_c\) as the position of the centroid of the template in the image frame. The filter states are the same as the vector of observations plus their first-order derivatives.

\begin{equation}
    \label{eq:5}
    \boldsymbol{X} = [x_c, y_c, O_w, O_h, \theta, \Dot{x_c}, \Dot{y_c}, \Dot{O_w}, \Dot{O_h}, \Dot{\theta}]^T
\end{equation}

Knowing the transition dynamics and states of the filter, the motion model of the system is then given by \eqref{eq:trans}. The matrix \(\boldsymbol{A} \in \mathbb{R}^{10 \times 10}\), shown in \eqref{eq:amat}, is the state transition matrix of the system and \(\boldsymbol{w}\) is a white noise random vector such that \(\boldsymbol{w} \sim \mathcal{N}(0, \boldsymbol{Q})\). \(\boldsymbol{Q} \in \mathbb{R}^{10 \times 10}\) is defined as the covariance matrix of the process noise \cite{x}. For the sake of notation, \(\mathbf{I}\) represents the identity matrix.

\begin{equation}
    \label{eq:trans}
    \boldsymbol{X}_t = \boldsymbol{A}_{t-1}\boldsymbol{X}_{t-1} + \boldsymbol{w}_{t-1}
\end{equation}

\begin{equation}
    \label{eq:amat}
    \boldsymbol{A} = \begin{pmatrix}
\mathbf{I}_{5 \times 5} &  \Delta t \mathbf{I}_{5 \times 5}  \\
\mathbf{0}_{5 \times 5} &  \mathbf{I}_{5 \times 5}
\end{pmatrix}_{10 \times 10}
\end{equation}

Likewise, the measurement model of the filter is given in \eqref{eq:obs}. \(\boldsymbol{H} = \mathbf{I}_{5 \times 10}\) is defined as the observation matrix and \(\boldsymbol{v}\) is a white noise random vector such that \(\boldsymbol{v} \sim \mathcal{N}(0, \boldsymbol{R})\). As for \(\boldsymbol{Q}\), here \(\boldsymbol{R} \in \mathbb{R}^{5 \times 5}\) is the covariance matrix of the observation noise. 

\begin{equation}
    \label{eq:obs}
    \boldsymbol{Y}_t = \boldsymbol{H}_{t}\boldsymbol{X}_{t} + \boldsymbol{v}_{t}
\end{equation}

With the motion and measurement models defined, it is possible to formulate the pose estimation process of the platform by giving the Kalman filter equations \eqref{eq:6}-\eqref{eq:12}. In this set of equations, \(\boldsymbol{P}\) is defined as the covariance matrix of the posterior estimate, \(\boldsymbol{Y}\) is the innovation vector, \(\boldsymbol{K}\) is the Kalman gain and \(\boldsymbol{S}\) is the covariance matrix of the innovation. Additionally, \(\boldsymbol{\check{X}}\) represents the predicted states and \(\boldsymbol{\hat{X}}\) the corrected states after a measurement update. 

The first two equations \eqref{eq:6}-\eqref{eq:7} are used in the prediction step, and give an estimate of the states \(\boldsymbol{X}\) at each time step regardless of whether an observation was obtained.

\begin{equation}
    \label{eq:6}
    \boldsymbol{\check{X}}_{t} = \boldsymbol{A}_{t} \boldsymbol{X}_{t-1} 
\end{equation}

\begin{equation}
    \label{eq:7}
    \boldsymbol{P}_{t} = \boldsymbol{A}_{t} \boldsymbol{P}_{t-1} \boldsymbol{A}_{t}^T + \boldsymbol{Q}_{t}
\end{equation}

When an observation is obtained by the detection module, the correction phase \eqref{eq:8}-\eqref{eq:12} is computed immediately after the prediction step. This step aims to correct the error in the estimations using an observation of the template in the current image frame at time \(t\).

\begin{equation}
    \label{eq:8}
    \boldsymbol{Y}_{t} = \boldsymbol{Z}_{t} - \boldsymbol{H}_{t} \boldsymbol{\check{X}}_{t}
\end{equation}

\begin{equation}
    \label{eq:9}
    \boldsymbol{S}_{t} = \boldsymbol{H}_{t} \boldsymbol{P}_{t} \boldsymbol{H}_{t}^T + \boldsymbol{R}_{t}
\end{equation}

\begin{equation}
    \label{eq:10}
    \boldsymbol{K}_{t} = \boldsymbol{P}_{t} \boldsymbol{H}_{t}^T \boldsymbol{S}_{t}^{-1}
\end{equation}

\begin{equation}
    \label{eq:11}
    \boldsymbol{\hat{X}}_{t} = \boldsymbol{\check{X}}_{t} + \boldsymbol{K}_{t} \boldsymbol{Y}_{t}
\end{equation}

\begin{equation}
    \label{eq:12}
    \boldsymbol{P}_{t} = (\boldsymbol{\mathbf{I}}_{10x10} - \boldsymbol{K}_{t} \boldsymbol{H}_{t}) \boldsymbol{P}_{t}
\end{equation}

\begin{figure}[ht]
    \centering
    \includegraphics[width=0.45\textwidth,height=0.45\textheight,keepaspectratio]{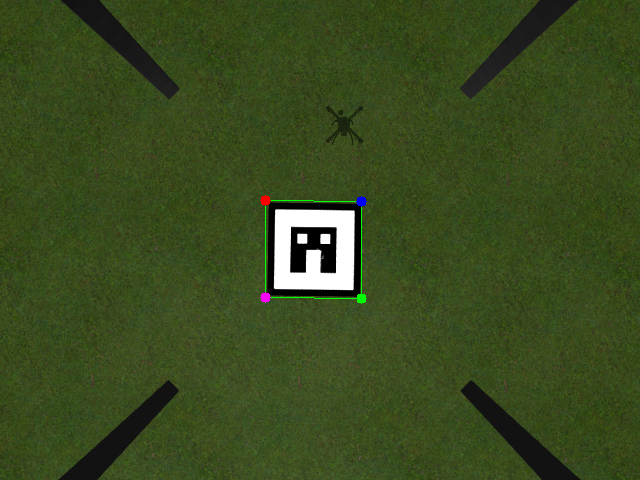}
    \captionsetup{justification=centering}
    \caption{Estimation of the platform position using the Kalman Filter vector of states \(\boldsymbol{X}\)}
    \label{fig:f3}
\end{figure}

The set of states produced by the Kalman filter can be applied in an IBVS module to control the landing procedure and to obtain the position of the landing platform in the current image frame, as shown in Fig. \ref{fig:f3}. These states are computed in the 2D image frame to reduce the computation carried out by the on-board computer of the UAV. Furthermore, since we are not estimating the relative pose between the vehicle and the landing platform, we are not feeding IMU data to the tracking module; this allows for the use of a linear Kalman filter and avoids the calculation of Jacobians at each time-step. Pseudo-code for the vision-based detection and tracking system is given in Algorithm \ref{alg:tracker}.

\begin{algorithm}
  \caption{Landing platform detector and tracker}
  \label{alg:tracker}
  \begin{algorithmic}[1]
    \Inputs{$\boldsymbol{\overline{H}} \in \mathbb{R}^{3 \times 3}$ Homography matrix\\
    $\boldsymbol{T} \in \mathbb{R}^{2}$ Dimensions of the template}
    \Initialization{$\boldsymbol{A} \in \mathbb{R}^{10 \times 10}$, \; $\boldsymbol{R} \in \mathbb{R}^{5 \times 5}$, \; $\boldsymbol{P} \in \mathbb{R}^{10 \times 10}$, \; $\boldsymbol{Q} \in \mathbb{R}^{10 \times 10}$, \;
    $\boldsymbol{K} \in \mathbb{R}^{5 \times 5}$, \;
    $\boldsymbol{S} \in \mathbb{R}^{10 \times 10}$, \;
    $\boldsymbol{H} \in \mathbb{R}^{5 \times 10}$, \;
    $\boldsymbol{Y} \in \mathbb{R}^{5}$, \; $\boldsymbol{X} \in \mathbb{R}^{10}$}
    \Outputs{$\boldsymbol{X} \in \mathbb{R}^{10}$ State vector}
    \Variables{$\boldsymbol{P_c} \in \mathbb{R}^{5}$ Corners and centroid of the template \\
    $\boldsymbol{Z} \in \mathbb{R}^{5}$ Observations of the template\\
    $O_w$ Width of the template in the image frame \\ 
    $O_h$ Height of the template in the image frame \\
    $\theta$ Estimated angle w.r.t X axis}
    \For{\textbf{each} frame}
        \For{$1, \dots, 5$}
            \State $\boldsymbol{P_c}_t$ $\gets$ computePoint($\boldsymbol{\overline{H}}, \boldsymbol{T}, point = i$) 
        \EndFor
        \State $\boldsymbol{P_c}_t$ $\gets$ sortCorners($\boldsymbol{P_c}_t$)
        \State $O_w, O_h \gets$ computeObjectDims($\boldsymbol{P_c}_t$)
        \State $\theta \gets$ computeAngle($\boldsymbol{P_c}_t$)
        \If{$\theta > 90 \deg$}
          \State $\theta_{new} \gets \theta/90$
          \State $\theta \gets \theta - \theta_{new} \times 90$
        \EndIf
        \State $\boldsymbol{Z} \gets$ $\{\boldsymbol{P_c}_t^{(i = 5)}, O_w, O_h, \theta\}$
        \State // KF Prediction step
        \State $\boldsymbol{\check{X}}_{t} \gets \boldsymbol{A}_{t} \times \boldsymbol{X}_{t-1} $ 
        \State $\boldsymbol{P}_{t} \gets \boldsymbol{A}_{t} \times  \boldsymbol{P}_{t-1} \times \boldsymbol{A}_{t}^T + \boldsymbol{Q}_{t}$ 
        \If{detection is valid}
            \State // KF Correction step
            \State $\boldsymbol{Y}_{t} \gets \boldsymbol{Z}_{t} - \boldsymbol{H}_{t} \times \boldsymbol{\check{X}}_{t}$ 
            \State $\boldsymbol{S}_{t} \gets \boldsymbol{H}_{t} \times \boldsymbol{P}_{t} \times \boldsymbol{H}_{t}^T + \boldsymbol{R}_{t}$ 
            \State $\boldsymbol{K}_{t} \gets \boldsymbol{P}_{t} \times \boldsymbol{H}_{t}^T \times \boldsymbol{S}_{t}^{-1}$ 
            \State $\boldsymbol{\hat{X}}_{t} \gets \boldsymbol{\check{X}}_{t} + \boldsymbol{K}_{t} \times \boldsymbol{Y}_{t}$
            \State $\boldsymbol{P}_{t} \gets (\boldsymbol{\mathbf{I}}_{10x10} - \boldsymbol{K}_{t} \times \boldsymbol{H}_{t}) \times \boldsymbol{P}_{t}$ 
        \EndIf
        \State $\boldsymbol{X}_{t} \gets \boldsymbol{\hat{X}}_{t}$
        \State \textbf{return} $\boldsymbol{X}_{t}$
    \EndFor
  \end{algorithmic}
\end{algorithm}


\section{Control System}

This section describes the PID-based controller used to autonomously land our rotorcraft. The IBVS controller uses the 2D output of the estimation module as a reference to compute position-velocity control signals to land the vehicle. These signals are sent to the native position-velocity loops implemented in the PX4 flight-stack, which transforms the positions into speeds and then converts them into thrust commands for the vehicle's engines, to guarantee correct control of the aircraft.


\subsection{PID-based controller}

In order to ensure that the aircraft moves towards the landing platform and lands on it, a control strategy is required. Autonomous landing of the vehicle is accomplished by feeding the position estimates of the template from the Kalman filter to a set of three PID-based controllers.

The IBVS PIDs will perform all the calculations in the current image frame. Setting a 2D image-based reference for the controller, and thus avoiding the need for expensive 3D reconstructions, increases the computational speed in on-board computers, allowing for real-time control over the approaches of the vehicle to the landing pad \cite{k}. High-rate controllers tend to be robust against sudden image changes, and with the Kalman filter output as the reference for control, the system is capable of tracking the landing platform even if it is abruptly moved out of the camera's field of view. 

Our approach uses a set of three PID-based controllers attached in a cascade in an outer loop, with the two native controllers already implemented in the Pixhawk flight stack. The controllers of the flight stack have a standard cascaded position-velocity loop, in which the outer position loop transforms the position inputs to velocity outputs and the velocity outputs are converted in the inner loop into thrust commands for the vehicle's propellers. The idea is to transform pixel coordinate errors into velocity commands and to let the inner controllers of the Flight Controller Unit (FCU) handle the thrust. 

We use a reference vector for the controllers \(\boldsymbol{S_p} = [\frac{I_w}{2}, \frac{I_h}{2}, 0]^T\), where \(\frac{I_w}{2}, \frac{I_h}{2}\) represent the center of the current image frame and zero corresponds to the desired angle between the aircraft and template measured with respect to the x axis of the image. These controllers command the \(x, y\) velocities of the rotorcraft, denoted as \(\Dot{x_a}, \Dot{y_a}\), to center the vehicle with respect to the template detected in the current image frame. The third controller modifies the yaw rate \(\Dot{\psi}\) of the aircraft in order to align it with the landing pad in the x axis. The error vector \(\boldsymbol{e}_t \in \mathbb{R}^3\) of the controllers at time \(t\) is given by \eqref{eq:13}.

\begin{equation}
    \label{eq:13}
    \boldsymbol{e}_t = \boldsymbol{S_p}_t - \boldsymbol{X}_{t}^{(i =  1:3)}
\end{equation}

\begin{figure}[ht]
    \centering
    \includegraphics[width=0.5\textwidth,height=0.5\textheight,keepaspectratio]{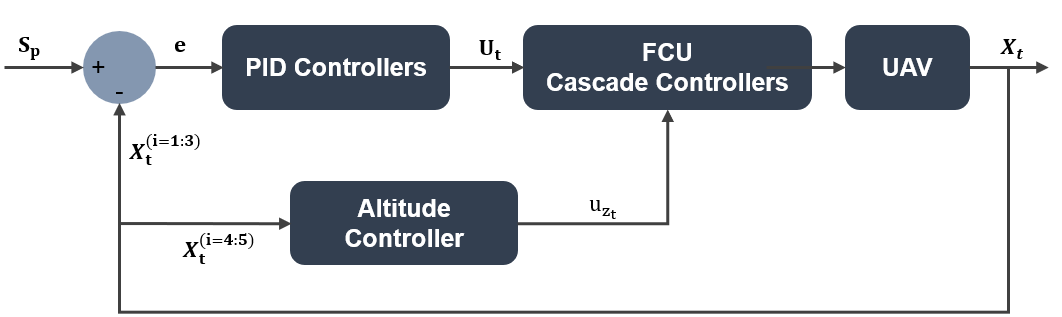}
    \captionsetup{justification=centering}
    \caption{The proposed control strategy attached to the native FCU controllers. The PID controllers block outputs the control signal \(\boldsymbol{U}_t\) with \(\Dot{x_a}, \Dot{y_a}, \Dot{\psi}\), whereas the altitude controller outputs \(u_{z_{t}}\) to control the descent of the vehicle.}
    \label{fig:f4}
\end{figure}

Fig. \ref{fig:f4} is a simplified representation of the three PID-based controllers and an altitude controller with an ON/OFF strategy to control the descent of the UAV. The error vector \(\boldsymbol{e}\) is used to feed the first three controllers and to produce a control effort \(\boldsymbol{U}_t \in \mathbb{R}^3\), which is delivered to the cascade controllers of the FCU. The output of the three PID controllers is provided by the vector in \eqref{eq:14}. Each PID controller was discretized using trapezoidal integration and derivation.

\begin{equation}
    \label{eq:14}
    \boldsymbol{U}_t = [\Dot{x_a}, \Dot{y_a}, \Dot{\psi}]^T
\end{equation}

The ON/OFF altitude controller in Fig. \ref{fig:f4} starts to land the aircraft whenever the difference between the height and width of the template estimates \(\boldsymbol{X}_{t}^{(i = 4:5)}\) tends to zero \eqref{eq:15}. \(u_{z_{t}}\) represents the output of the altitude controller, \(Z_p\) is the current height in meters of the vehicle and \(Z_f\) is a descent constant. This descent condition guarantees that the aircraft will land only if the template dimensions form a square, which is the actual shape of the landing platform.

\begin{equation}
    \label{eq:15}
    u_{z_{t}} =
    \begin{cases}
      Z_p - Z_f, & \text{if}\ |O_w - O_h| < 5 \\
      Z_p, & \text{otherwise}
    \end{cases}
\end{equation}

Acquiring feedback from the vision-based module closes the visual servoing control loop and allows for the implementation of an on-board end-to-end control strategy for a UAV. Algorithm \ref{alg:controller} shows pseudo-code for the controller pipeline for the rotorcraft. 


\begin{algorithm}
  \caption{Landing controller}
  \label{alg:controller}
  \begin{algorithmic}[1]
    \Inputs{$\boldsymbol{X} \in \mathbb{R}^{10}$ State vector \\
    $I_w$ image Width \\
    $I_h$ image Height}
    \Init{$\boldsymbol{PID_{x_a}} \in \mathbb{R}^3$ PID parameters for $\Dot{x_a}$ \\
    $\boldsymbol{PID_{y_a}} \in \mathbb{R}^3$ PID parameters for $\Dot{y_a}$ \\
    $\boldsymbol{PID_{\psi}} \in \mathbb{R}^3$ PID parameters for $\Dot{\psi}$ \\
    $\boldsymbol{S_p} \in \mathbb{R}^3$ \(x, y,\) and \( \theta\) setpoints \\
    $Z_p$ initial height of the vehicle \\
    $Z_f$ descent factor}
    \Outputs{$\boldsymbol{U} \in \mathbb{R}^3$ PID control efforts \\
    $u_z$ ON/OFF altitude controller output }
    \For{\textbf{each} state vector X}
        \State $\boldsymbol{e}_t \gets \boldsymbol{S_p}_t - \boldsymbol{X}_{t}^{(i = 1:3)}$
        \State $errorSize \gets abs(\boldsymbol{X}_{t}^{(i = 4)} - \boldsymbol{X}_{t}^{(i = 5)})$
        \State // Update Z position
        \If{$errorSize < 5 \text{ and } Z_p > 0.2$} 
            \State $u_{z_{t}} \gets Z_p - Z_f$
        \Else
            \State $u_{z_{t}} \gets Z_p$
        \EndIf
        \State // Land if the vehicle is at 0.2 meters from the ground
        \If{$u_{z_{t}} <= 0.2 \text{ and } (\boldsymbol{e}_{t}^{(i = 1)}, \boldsymbol{e}_{t}^{(i = 2)}) < 20$}
            \State land()
            \State $u_{z_{t}} \gets 0$
        \EndIf
        \State $  Z_p \gets u_{z_{t}}$ 
        \For{$i = 1, \dots, 3$}
            \State $ \boldsymbol{U}^{(i)} \gets computePID(\boldsymbol{S_p}_{t}^{(i)}, \boldsymbol{e}^{(i)}, \boldsymbol{X}^{(i)})$ 
        \EndFor
        \State \textbf{return} $\boldsymbol{U}_t, u_{z_{t}}$
    \EndFor
  \end{algorithmic}
\end{algorithm}


\section{Simulation results}

This section describes the experiments carried out to assess the different modules of the autonomous landing system using a Gazebo-based simulation. We provide an open-source implementation of our system in Github\footnote[1]{ \url{https://github.com/MikeS96/autonomous_landing_uav}}.

\begin{figure}[t]
  \centering
  \begin{subfigure}{.48\columnwidth}
    \centering
    \includegraphics[width=\linewidth,]{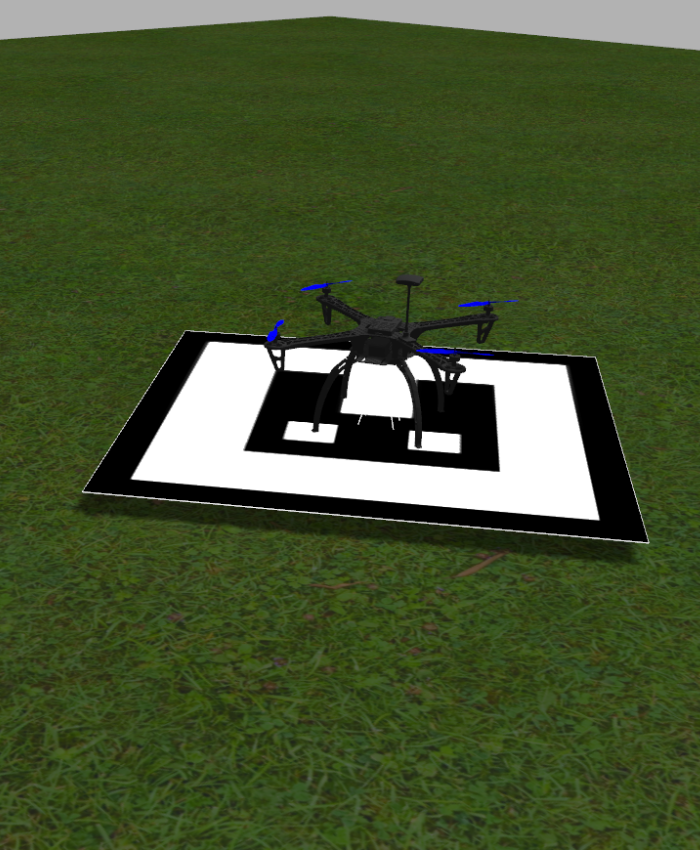}
    \caption{}
  \end{subfigure}%
  \hfill
  \begin{subfigure}{.48\columnwidth}
    \centering
    \includegraphics[width=\linewidth]{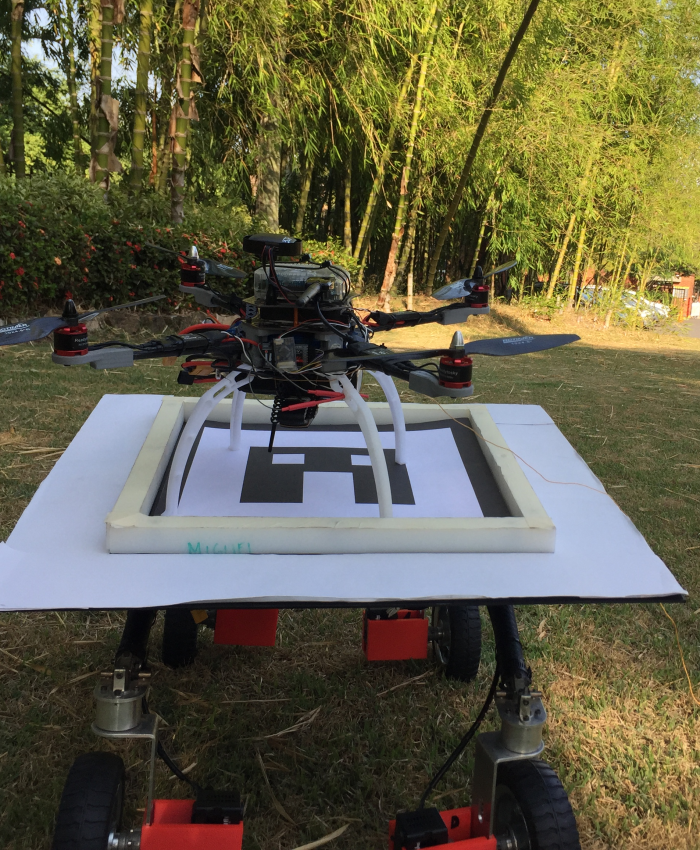}
    \caption{}
  \end{subfigure}%
  \hfill
  \captionsetup{justification=centering}
  \caption{DJI F450 quad-rotor in the landing pad: (a) Dron in the Gazebo-based simulation; (b) Customized dron with a Pixhawk FCU and Odroid XU4 in field trial.}
  \label{fig:f10}
\end{figure}

The system was simulated using the SITL provided by PX4, which runs the Pixhawk flight stack in a Gazebo-based environment. Our implementation relies on the SITL simulation environment presented in \cite{1x}, where the PX4 on SITL is connected via UDP with an offboard API (ROS), ground station and the gazebo simulator. To obtain accurate results, a custom model of a DJI F450 quad-rotor was implemented to mimic the dynamics and physics involved in a real-world model, as shown in Fig. \ref{fig:f10} (a). All the perception and control pipelines of the system, shown in Fig. \ref{fig:f1}, were implemented in the Robot Operating System (ROS). In addition, a custom Gazebo-world with a landing platform was used to rigorously assess the performance of both the vision and control module.

\begin{figure*}[ht]
  \centering
  \includegraphics[height=230px, keepaspectratio]{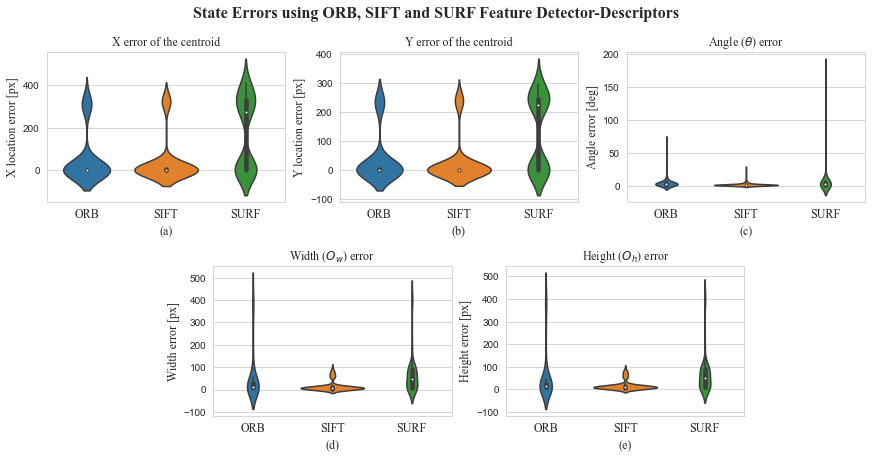}
  \caption{Estimation errors for the different descriptor-detector algorithms used for generating state observations (\(\boldsymbol{Z}\)): ORB to the left, SIFT in the middle and SURF to the right, of each sub-figure respectively. (a) centroid's \(X\) error; (b) centroid's \(Y\) error; (c) Orientation \(\theta\) error; (d) width \(O_w\) error; (e) height \(O_h\) error.}
  \label{fig:f6}
\end{figure*}

\subsection{Vision module}

The assessment of the detection and tracking system was carried out using three different detector-descriptor algorithms, which are efficient to compute and, orientation and scale invariant \cite{1w}: ORB, SIFT and SURF. After extracting landing pad detections from the captured aerial images as explained in Section \ref{section:fbod}, the Kalman filter was used to estimate the state of the landing pad. Our evaluation procedure demonstrates the improvements obtained by our tracking module compared with plain detection. All the detector-descriptors were tested with the aircraft hovering at a height of \(3.5\) meters above the landing pad.

The RANSAC algorithm was used to compute the homography matrix \(\boldsymbol{\overline{H}}\). Both SIFT and SURF used the Manhattan distance to compute the matches between descriptors, whereas ORB employed the Hamming distance. Figure \ref{fig:f6} illustrates the results of the three algorithms. The violin plots show the error between the observations \(\boldsymbol{Z}\) and the ground truth of the platform. These plots show the distribution of the error for the five observed states, with the median error represented as a white dot, the interquartile range as a broad black bar in the center of the violin, and the lower/upper adjacent values as a thin line.

It can be seen from Fig. \ref{fig:f6} (a) and (b) that the centroid coordinates \(x,y\) of the landing platform show similar behavior for the ORB and SIFT detectors, with a median value close to zero. In contrast, SURF has more dispersion in its error distribution and a median of above \(200\) pixels. The best detector for the centroid coordinates is SIFT, as it gives a more uniform distribution compared with ORB and SURF, and most of the error values are clustered close to zero.

Figure \ref{fig:f6}(c) presents the error in the angle \(\theta\), and it can be observed that the SIFT detector gives better performance than the other two detectors. The errors in the width \(O_w\) and height \(O_h\) can be seen in Fig. \ref{fig:f6}(d) and (e), respectively. From this figure, it can be seen that the three detectors have very similar behavior for both variables, although SIFT outperforms ORB and SURF with an error distribution close to zero and few outliers.

The SIFT detector-descriptor is better than the other detectors for all observations \(\boldsymbol{Z}\). Although ORB shows similar behavior to SIFT for the first three states, it has a large set of outliers for the last two states, while SURF gives the worst performance throughout the observation space. 

\begin{table}[h]
\captionsetup{justification=centering, labelsep=newline}
\caption{Comparison between plain SIFT detector and SIFT detector with Kalman filter}
\label{tab:detector_kf}
\centering
\begin{tabular}{@{}lcccc@{}}
\toprule
\multicolumn{1}{c}{\multirow{2}{*}{States}} & \multicolumn{2}{c}{Using SIFT only} & \multicolumn{2}{c}{SIFT with Kalman Filter}\\ \cmidrule(l){2-5} 
\multicolumn{1}{c}{}                        & Average & \begin{tabular}[c]{@{}c@{}}Standard\\ deviation\end{tabular} & Average & \begin{tabular}[c]{@{}c@{}}Standard\\ deviation\end{tabular} \\ \midrule
Centroid X [px]  & 41.93  & 105.12 & 2.74  & 4.71  \\
Centroid Y [px]  & 30.16  & 78.55  & 1.12  & 2.91  \\
Angle [deg] & 1.81   & 2.69   & 1.31  & 2.02  \\
Width [px]  & 15.98  & 22.40  & 7.00  & 9.71  \\
Height [px]  & 18.08  & 20.43  & 9.36  & 6.20  \\ 
\bottomrule
\end{tabular}
\end{table}

Finally, Table \ref{tab:detector_kf} shows the average and standard deviation in the errors in pixels between the SIFT detector and the Kalman filter attached to the SIFT detector. The results demonstrate that all of the observations are substantially improved with the Kalman filter, reducing the average error to almost zero and decreasing the standard deviation of each state. This analysis leads to the conclusion that the detection and tracking pipeline can accurately track the landing platform with a SIFT detector and a linear Kalman filter to facilitate the computations in the on-board computers of a small UAV.

\begin{figure*}
  \centering
  \includegraphics[height=230px, keepaspectratio]{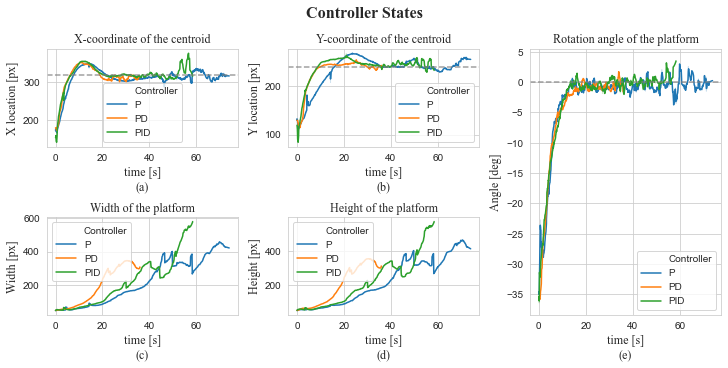}
  \caption{Output of the P, PD and PID controllers for each state in \(\boldsymbol{X}_t^{(i = 1:5)}\): (a) \(X\) coordinate of the centroid; (b) \(Y\) coordinate of the centroid; (c) width of the platform \(O_w\); (d) height of the platform \(O_h\); (e) heading \(\theta\).}
  \label{fig:f7}
\end{figure*}

\begin{figure*}
  \centering
  \includegraphics[height=160px, keepaspectratio]{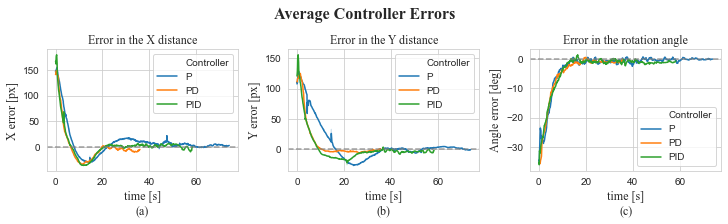}
  \caption{Error in the P, PD and PID controllers for \(\boldsymbol{X}_t^{(i = 1:3)}\): (a) \(X\) error for the centroid; (b) \(Y\) error for the centroid; (c) error in the heading \(\theta\).}
  \label{fig:f8}
\end{figure*}

\subsection{Controller}

To assess the performance of the IBVS control system, three PID-based strategies were implemented. Various tests were carried out with P, PD and PID controllers to determine which was optimal for the landing procedure. For each control strategy, the gains were tweaked in a gazebo-based simulation until the most stable parameters were found for each controller. Using the best gains, five landing trials were conducted in a custom Gazebo environment, and the results were averaged. The image size was set to $640 \times 320$ pixels, and the SIFT detector-descriptor was employed.

Fig. \ref{fig:f7} presents the output of the three controllers for each state \(\boldsymbol{X}_t^{(i = 1:5)}\). The first two figures (Fig. \ref{fig:f7}(a) and (b)) correspond to the \(x,y\) centroid of the landing platform. It can be seen that the three controllers were capable of tracking the 2D reference provided by the vision module and to center the vehicle on the pad. However, the P strategy (blue) operated more slowly than the PD (orange) and PID (green) strategies, which tended to land the aircraft faster.

In a similar fashion, all of the controllers were shown to be capable of aligning the heading of the vehicle with the landing platform, as shown in Fig. \ref{fig:f7}(e). The estimated width \(O_w\) and height \(O_h\) of the landing pad, as illustrated in Fig. \ref{fig:f7}(c) and (d), have a tendency to increase as the altitude controller starts the vehicle's descent. This effect is due to the landing platform becoming bigger in the current image frame as the height of the aircraft decreases.


\begin{table}[]
\centering
\captionsetup{justification=centering, labelsep=newline}
\caption{Controller errors in the landing process}
\label{tab:control-error}
\begin{tabular}{cllll}
\toprule
Controller & \multicolumn{1}{c}{} & \multicolumn{1}{c}{\begin{tabular}[c]{@{}c@{}}Centroid X\\ {[}px{]}\end{tabular}} & \multicolumn{1}{c}{\begin{tabular}[c]{@{}c@{}}Centroid Y\\ {[}px{]}\end{tabular}} & \multicolumn{1}{c}{\begin{tabular}[c]{@{}c@{}}Angle\\ {[}deg{]}\end{tabular}} \\ \midrule
P    & RMSE                & 30.3725  & 34.0250  & 6.7023  \\
     & Standard deviation  & 28.9982  & 32.9310  & 6.1946  \\ \midrule
PD   & RMSE                & 41.8702  & 40.2214  & 11.3106  \\
     & Standard deviation  & 41.5917  & 37.2056  & 9.4058  \\ \midrule
PID  & RMSE                & 31.8581  & 29.0585  & 7.7358  \\
     & Standard deviation  & 31.6203  & 29.0283  & 6.9005  \\ \midrule
\end{tabular}
\end{table}

Although all of the controllers were capable of landing the aircraft, in order to perform a thorough assessment we present, the error for each controller for the states \(\boldsymbol{X}_t^{(i = 1:3)}\) in Table \ref{tab:control-error}. From this table, it can be seen that the RMSE and standard deviation (in pixels) for each controller are strikingly similar for the three states under consideration. The P and PID controllers gave better numerical results than the PD controller. However, this behavior was due to the landing speed of the PD controller; since it is capable of landing more quickly, there are fewer samples to compute the RMSE. The PD controller landed in approximately \(36\) seconds, around \(25\) seconds faster than the PID controller.


Fig. \ref{fig:f8} complements the information in Table \ref{tab:control-error} by presenting the dynamic behavior of the error in the first three states \(\boldsymbol{X}_t^{(i = 1:3)}\) for each controller while the aircraft is landing. As mentioned above, the P controller is slower than the other two controllers. PD tends to be a faster strategy and has fewer overshoots in its dynamic behavior. The  performance of PID seems to be between those of the other two controllers.

\begin{figure*}
  \centering
  \includegraphics[height=230px, keepaspectratio]{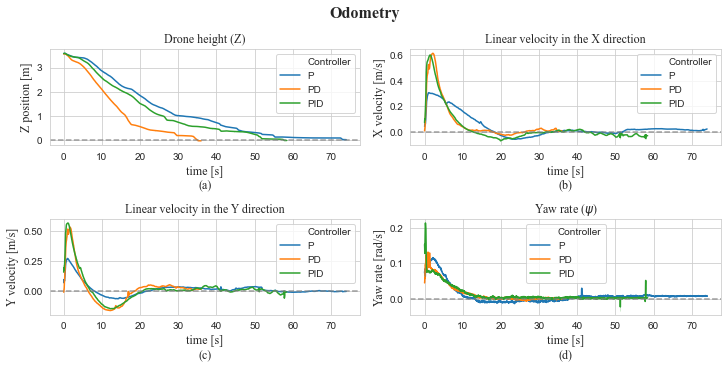}
  \caption{Odometry of the vehicle during the landing process: (a) height of the vehicle \(Z_p\); (b) \(X\) velocity \(\Dot{X_a}\); (c) \(Y\) velocity \(\Dot{Y_a}\), (d) yaw rate \(\Dot{\psi}\).}
  \label{fig:f9}
\end{figure*}

The odometry of the vehicle is presented in Fig. \ref{fig:f9} for four different variables for each controller. The first plot in Fig. \ref{fig:f9}(a), shows how the altitude of the vehicle is reduced to zero for each controller. Both of the linear velocities of the aircraft \(\Dot{x_a},\Dot{y_a}\) undergo substantial variation at the beginning of the tests, as shown in Fig. \ref{fig:f9}(b) and (c), but when the vehicle is centered with respect to the landing platform, the linear speeds tend to zero. Likewise, the yaw rate \(\Dot{\psi}\) shown in Fig. \ref{fig:f9}(e) behaves as expected for the three controllers: its magnitude reduced to zero, which means that the vehicle is correctly aligned with the landing platform. 

The position and heading errors between the landing platform and the aircraft were computed for the different trials, and are shown in Table \ref{tab:landing-error}. It can be seen from the table that the average error in the \(x, y\) coordinates is less than \(3.0\) centimeters for all the control strategies. Similarly, the error in the angle \(\theta\) between the vehicle and the platform is less than \(1.2\) degrees. This demonstrates that all of the controllers are capable of achieving a precision landing of the aircraft with small errors over different trials, confirming the efficiency of the vision-based system with various control strategies. 

\begin{table}[]
\centering
\captionsetup{justification=centering, labelsep=newline}
\caption{Errors in the landing simulation}
\label{tab:landing-error}
\begin{tabular}{@{}cllll@{}}
\toprule
Controller & \multicolumn{1}{c}{} & \multicolumn{1}{c}{\begin{tabular}[c]{@{}c@{}}Centroid X\\ {[}m{]}\end{tabular}} & \multicolumn{1}{c}{\begin{tabular}[c]{@{}c@{}}Centroid Y\\ {[}m{]}\end{tabular}} & \multicolumn{1}{c}{\begin{tabular}[c]{@{}c@{}}Angle\\ {[}deg{]}\end{tabular}} \\ \midrule
P    & Average              & 0.0244  & 0.0294  & 0.6531  \\
     & Standard deviation   & 0.0079  & 0.0178  & 0.3770  \\ \midrule
PD   & Average              & 0.0178  & 0.0274  & 0.7563  \\
     & Standard deviation   & 0.0194  & 0.0155  & 0.8444  \\ \midrule
PID  & Average              & 0.0288  & 0.0232  & 1.1115  \\
     & Standard deviation   & 0.0261  & 0.0253  & 0.5323  \\ \bottomrule
\end{tabular}
\end{table}

Although all of the controllers were capable of accurately landing the vehicle on the landing platform, the best performance was shown by the PD and PID controllers, as these had more stable responses and lower variations in the different attempts. Although the PD strategy is less accurate than the PID controller, it is the preferred option due to its speed in landing the aircraft.

\begin{figure*}
  \centering
  \includegraphics[height=160px, keepaspectratio]{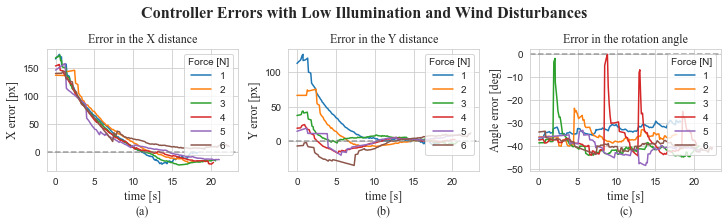}
  \caption{Error in the PD controller for \(\boldsymbol{X}_t^{(i = 1:3)}\) with low illumination and different wind disturbances: (a) \(X\) error for the centroid; (b) \(Y\) error for the centroid; (c) error in the heading \(\theta\).}
  \label{fig:error}
\end{figure*}

To assess the robustness of the PD controller under low light conditions and different wind disturbance, Fig. \ref{fig:error} presents the errors obtained over different trials while the aircraft is landing. The error in \(X\) illustrated in Fig. \ref{fig:error}(a) shows how the aircraft is capable to minimize it towards zero with different wind conditions. Similarly, the error in \(Y\) presented in \ref{fig:error}(b) demonstrates a similar behavior as \ref{fig:error}(a) where the error is minimized, nevertheless, with bigger wind disturbances the aircraft is prone to experience an overdamped response rather than underdamped as Fig. \ref{fig:f8} demonstrated. Finally, the angle \(\theta\) is considerably affected by the wind disturbances in \ref{fig:error}(c) as the vehicle is not capable to align itself with respect to the landing platform. However, the vehicle was capable to land in all tests, validating the effectiveness of our method while landing with unideal conditions.

\section{Experimental results}

This section presents the results obtained in real-world tests using a DJI F450 in an autonomous landing sequence.

To thoroughly assess the performance of the autonomous landing system, a custom DJI F450 with an Odroid XU4 on-board computer and Pixhawk 1, as shown in Fig. \ref{fig:f10} (b), was used to test the developed framework. Due to the limited computational resources of the Odroid, the PD controller was employed, as this was the fastest method of landing the vehicle, and the result of three landing trials were averaged to evaluate the system. The size of the image was also reduced to 320 x 240 pixels to obtain a frame rate of \(15\) FPS and to ensure system convergence. The SIFT feature detector-descriptor was used (based on the simulation results) to carry out these tests.

To bridge the algorithms developed during the simulation phase with the real-world, it was necessary to unplug the SITL component. This was achieved by connecting the Pixhawk FCU to the on-board computer and launching all the nodes developed in ROS. This process guaranteed that the system was connected with the physical FCU, bypassing the need for the SITL component. The detection and control pipeline will therefore operate directly in the custom rotor-craft, enabling it to carry out autonomous landing maneuvers. All the parameters used during the simulation where transferred to the aircraft without \textit{finetuning} to demonstrate that the use of simple vision and control models allow for zero-shot domain transfer.

\begin{figure*}[ht]
  \centering
  \includegraphics[height=230px, keepaspectratio]{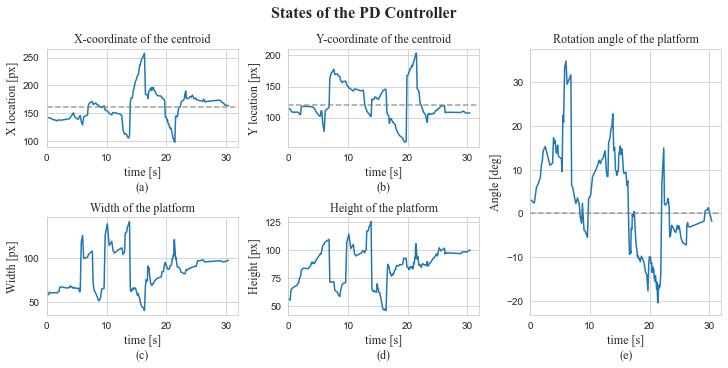}
  \caption{Output of the PD controller for each state in \(\boldsymbol{X}_t^{(i = 1:5)}\), as tested with a DJI F450: (a) X-coordinate of the centroid; (b) Y-coordinate of the centroid; (c) width of the platform \(O_w\); (d) height of the platform \(O_h\); (e) heading \(\theta\).}
  \label{fig:f11}
\end{figure*}

Fig. \ref{fig:f11} presents the results obtained with the PD controller for each state \(\boldsymbol{X}_t^{(i = 1:5)}\). As expected, the system is capable of landing the rotor-craft on the landing platform within approximately \(35\) seconds. The real-world system displays more spiky behavior than the simulated vehicle (Fig. \ref{fig:f7} (orange)); however, as the test advances, the response of the controller stabilizes, guaranteeing the appropriate landing of the UAV.

Comparably, the RMSE of the controller during the landing procedure was also assessed and presented in Table \ref{tab:control-error-real}. This error was computed over the three landing trials conducted with the real-world rotor-craft while the vehicle was trying landing. Altogether, it is possible to appraise that the vehicle maintains strikingly similar values of RMSE for the variables \(X, Y, \theta\) when compared with the RMSE presented for the simulation in Table \ref{tab:control-error}. In fact, the RMSE is slightly reduced within the real-world landing trials. The plots of these errors are unshown as their dynamic behavior is similar as the ones presented in Fig. \ref{fig:f8} (orange).

\begin{table}[]
\centering
\captionsetup{justification=centering, labelsep=newline}
\caption{Controller errors in experimental landing process}
\label{tab:control-error-real}
\begin{tabular}{cllll}
\toprule
Controller & \multicolumn{1}{c}{} & \multicolumn{1}{c}{\begin{tabular}[c]{@{}c@{}}Centroid X\\ {[}px{]}\end{tabular}} & \multicolumn{1}{c}{\begin{tabular}[c]{@{}c@{}}Centroid Y\\ {[}px{]}\end{tabular}} & \multicolumn{1}{c}{\begin{tabular}[c]{@{}c@{}}Angle\\ {[}deg{]}\end{tabular}} \\ \midrule

PD   & RMSE                & 39.7460  & 36.4111  & 8.7761  \\
     & Standard deviation  & 39.7583  & 31.5632  & 8.7724  \\ \midrule

\end{tabular}
\end{table}


Finally, to complete the assessment process, Table \ref{tab:landing-error-exp} \cite{y} presents the position error between the vehicle and the landing platform. This error was computed as the distance from the center of the pad to the center of the rotor-craft once it had landed. It can be seen that the average value is less than \(16\) centimeters. Compared to the results in Table \ref{tab:landing-error}, the error in the real-world implementation of the PD controller is around five times that of the simulation. Although these results seem undesirable, the rotor-craft is capable of precisely landing on the desired platform and accomplishing the autonomous landing task, as expected from the simulation results.

\begin{table}[]
\centering
\captionsetup{justification=centering, labelsep=newline}
\caption{Experimental landing errors measured from the center of the pad to the center of the vehicle}
\label{tab:landing-error-exp}
\begin{tabular}{@{}cllll@{}}
\toprule
Controller & \multicolumn{1}{c}{} & \multicolumn{1}{c}{\begin{tabular}[c]{@{}c@{}}Centroid X\\ {[}m{]}\end{tabular}} & \multicolumn{1}{c}{\begin{tabular}[c]{@{}c@{}}Centroid Y\\ {[}m{]}\end{tabular}} \\ \midrule
PD   & Average              & 0.1314  & 0.1592 \\
     & Standard deviation   & 0.1041  & 0.1344  \\ \bottomrule

\end{tabular}
\end{table}

\section{Conclusion}
This paper presents a SITL approach to developing a monocular image-based autonomous landing system for quadcopter drones. The proposed method and system, which integrates ROS, Gazebo and PX4's SITL tools, enables users to not only endow quadcopters with low-cost vision-based autonomous landing capabilities, but also to fine-tune all the parameters of a potentially dangerous device in a safe simulated environment. With minimal modifications, both the vision and control modules developed in our simulated environment, were successfully validated in a physical DJI F450 with an Odroid XU4 on-board computer and a Pixhawk 1 flight controller.


%



\section*{Acknowledgment}

This work was funded by Universidad Autónoma de Occidente (UAO), project 17INTER-297. The authors would like to thank the Robotics and Autonomous Systems (RAS) research incubator and UAO's Vicerrectoría de Investigaciones, Innovación y Emprendimiento for their support.

\ifCLASSOPTIONcaptionsoff
  \newpage
\fi

\end{document}